\pdfoutput=1

\documentclass[11pt]{article}

\usepackage{acl}

\usepackage{times}
\usepackage{latexsym}
\usepackage{expex}
\usepackage{graphicx}
\usepackage{enumitem}
\usepackage[T1]{fontenc}

\usepackage[utf8]{inputenc}
\usepackage{microtype}
\usepackage{booktabs}
\usepackage{inconsolata}

\lingset{aboveexskip=1ex,belowexskip=1ex}

\title{Implicit causality in GPT-2: a case study}

\author{Hien Huynh \and Tomas O. Lentz  \and Emiel van Miltenburg \\
Department of Communication and Cognition\\
Tilburg University\\
\texttt{\{m.h.huynh\_1, t.o.lentz, C.W.J.vanMiltenburg\}@tilburguniversity.edu}\\}

\begin{document}
\maketitle
\begin{abstract}
This case study investigates the extent to which a language model (GPT-2) is able to capture native speakers' intuitions about \emph{implicit causality} in a sentence completion task. We first reproduce earlier results (showing lower surprisal values for pronouns that are congruent with either the subject or object, depending on which one corresponds to the implicit causality bias of the verb), and then examine the effects of gender and verb frequency on model performance.
Our second study examines the reasoning ability of GPT-2: is the model able to produce more sensible motivations for why the subject \textsc{verb}ed the object if the verbs have stronger causality biases? We also developed a methodology to avoid human raters being biased by obscenities and disfluencies generated by the model.
\end{abstract}

\section{Introduction}
This paper is a case study, highlighting different ways to analyse the linguistic abilities of a language model, with respect to an established linguistic phenomenon, namely Implicit causality (IC) bias \cite{hartsthorne2014what}. Speakers associate either the subject or the object of a verb with the cause of the state or event described by that verb. For example, the verb \textit{frighten} is a \emph{subject-biased} verb because native speakers of English tend to see the subject as the cause of the frightening event. Thus, given a main clause like in (\ref{ex1:a}), participants in a sentence completion task would tend to provide a reason referring to the subject, as in (\ref{ex1:b}).

\newcommand{\lb}[1]{\upshape[\textsubscript{\textsc{#1}}}
\newcommand{\rb}{\upshape]}
\lingset{labeloffset=5px,textoffset=5px}
\pex %
\a \label{ex1:a}\lb{main clause} John scared Mary because \ldots\rb 
\a \label{ex1:b}\lb{reason} he put on a Halloween costume.\rb
\xe

Earlier work by \citet{upadhye-etal-2020-predicting} and \citet{davis-van-schijndel-2020-discourse,davis-van-schijndel-2021-uncovering} investigated the extent to which language models are able to capture native speakers' IC biases. This paper aims to reproduce some of their earlier results, using GPT-2 \cite{radford2019language} as an example. Using the same sentence completion task as \citet{davis-van-schijndel-2020-discourse}, we further investigate how bias, subject gender, and verb frequency influence the behavior of this model (\S\ref{sec:study1}). Next, we will more thoroughly assess the quality of the completions generated by GPT-2 (\S\ref{sec:study2}), asking: does the model's performance hold up to further scrutiny?

\paragraph{Why GPT-2?} Although it is neither the most recent, nor the best performing open-source language model around (see \citealt{black-etal-2022-gpt} for alternatives), GPT-2 is still a very popular choice for many researchers and practitioners.\footnote{See \url{https://huggingface.co/gpt2} for statistics.} This popularity is at least partly due to its size, as the model can be run and fine-tuned on consumer hardware.\footnote{The model's popularity means that studying its capabilities and limitations may be more impactful (at least in the short term) than studying the capabilities and limitations of larger but less accessible systems.} For us, GPT-2 offers the right balance of complexity and efficiency; as shown by \citet{upadhye-etal-2020-predicting}, its outputs are good enough to have a meaningful discussion about the assessment of language model performance, without requiring a large (and expensive) computational infrastructure.

\paragraph{Contributions} Next to the value of reproducing earlier work, and providing further details on the IC-related behaviour of GPT-2, the main innovation of this paper is the controlled assessment of output quality. We took great care to separate issues with the fluency and offensiveness of the output from the content of the generated continuations.\footnote{See Appendix~\ref{app:limitations} for the limitations of this study.} This not only makes the task less harmful to our participants, but it also increases their focus on our construct of interest: the reasoning abilities of language models.

\section{Data}\label{sec:prompts}
We present two studies. Study 1 investigates next-word surprisal values, and Study 2 looks at continuations that are generated based on a prompt. For both our studies, we provide the model with input sentences of the following form: 

\ex\label{ex:prompt}
\textit{\textsc{Subject} \textsc{verb}-ed \textsc{object} because \ldots}.
\xe

The verbs are derived from a list of 246 IC verbs compiled by
\citet{ferstl2011implicit}, who also provide a \emph{bias score} for each of these verbs. This score is derived from a human experiment, where participants were asked to complete sentences like (\ref{ex1:a}). The human bias scores range from -100 (i.e., all valid continuations produced by respondents in Ferstl et al.’s experiment uniquely referred to objects of the preceding clauses) to 100 (i.e., all valid continuations referred to subjects of the preceding clauses).

The subjects and objects are provided by \citet{davis-van-schijndel-2020-discourse}, who produced a list of 14 noun pairs that are grammatically male and female (e.g. \emph{man, woman} or \emph{brother, sister}). A combination of the nouns and verbs, our set of stimuli consists of 6888 examples (246 verbs × 14 pairs of gender-mismatched nouns × 2 subject genders).

\section{Study 1: IC-bias and pronoun use}\label{sec:study1}
First, we investigate whether verbs in GPT-2 possess the same subject/object bias as in the human experiment described above. 

\paragraph{Set-up} To test the hypothesis, we use the approach from \citet{davis-van-schijndel-2021-uncovering} and checked for each prompt whether the model assigned a lower surprisal to a male or female pronoun (i.e.\ \emph{he} or \emph{she}).\footnote{Next-word surprisal is estimated for a target upcoming word by taking the inverse log of this word's probability: $\textit{surprisal}(w_{t}) = -\textit{log} P(w_{t} | w_{1} \ldots w_{t-1})$} If GPT-2 captures the IC bias, then subject-biased verbs with a female subject should lead the model to produce lower surprisal values for \emph{she}. Since each noun pair is used in both orders (either a male or a female noun in subject position), we have a perfectly balanced dataset.

\begin{figure}
    \centering
    \includegraphics[width=\linewidth]{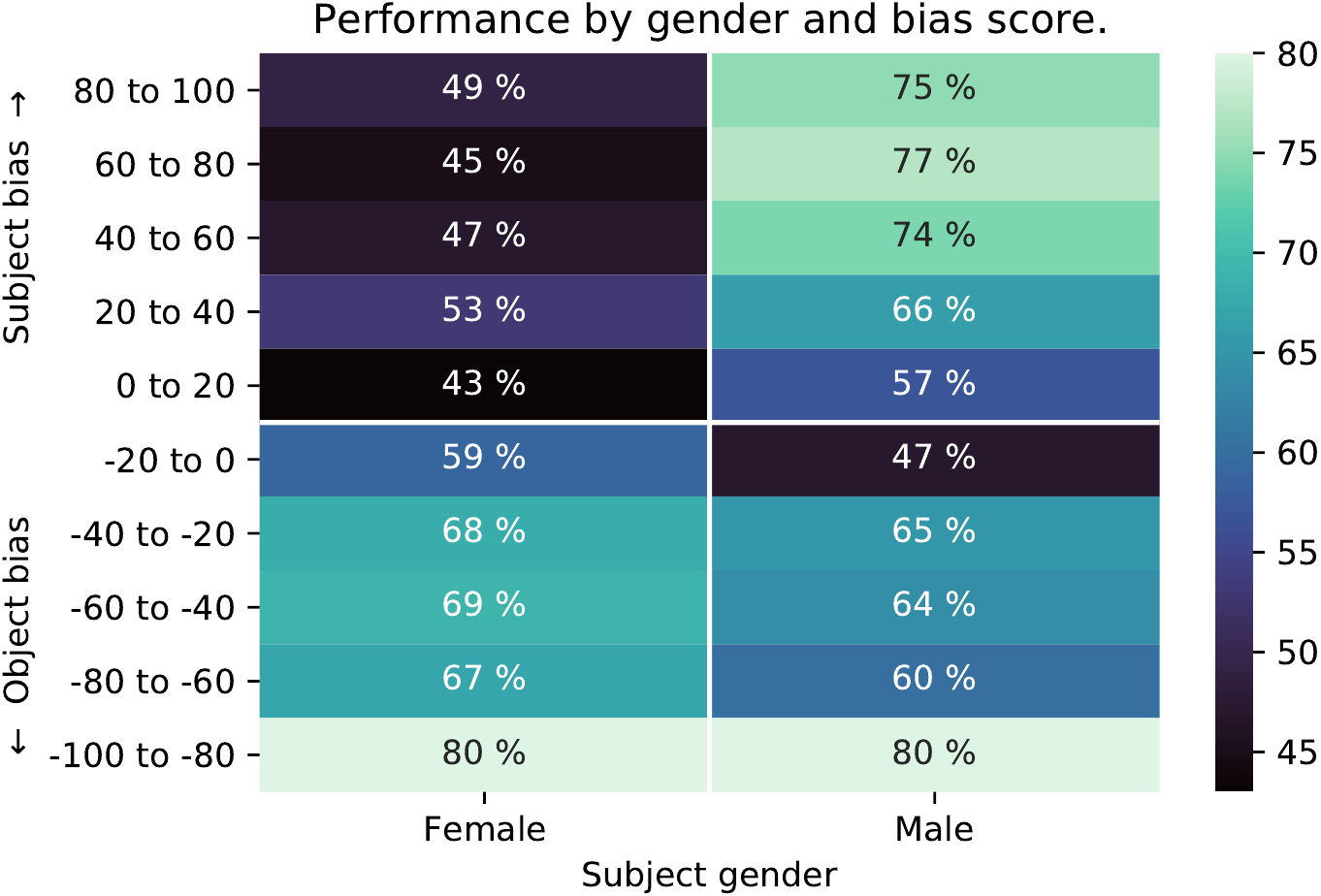}
    \caption{Heatmap table showing the percentage of outputs matching the subject or object bias of the verb, with scores separated by subject gender.}
    \label{fig:heatmaptable}
\end{figure}

\paragraph{Results} 
Figure~\ref{fig:heatmaptable} shows the results split by subject gender and bias scores from \citet{ferstl2011implicit}. GPT-2 generally picks up on the subject or object bias of the verb. The gender produces more subject-based explanations if the verb's IC bias is more subject-biased. There is only one exception: Performance for sentences with both subject-biased verbs and female subjects is at chance level for all bias scores.

\textit{IC Bias and Gender.} We used the lme4 \cite{lme4} and lmerTest package in R \cite{JSSv082i13} to carry out a mixed effects regression analysis of the relationship between the bias scores and GPT-2's subject-preference scores, corresponding to the difference between the surprisal for the object-congruent pronoun and the surprisal of the subject-congruent pronoun. Details about our statistical analyses are provided in Appendix~\ref{app:study1-statistics}. There were significant fixed effects of human bias score ($b = 0.003, \textrm{SE} = 0.0001, p < .001$) and of the interaction between human bias score and subject gender ($b = 0.0014, \textrm{SE} = 0.0002, p < .001$). This confirms our observations from Figure~\ref{fig:heatmaptable}. The effect of subject gender was found to be not significant ($b = -0.0103, \textrm{SE} = 0.012, p = .374$).\footnote{In addition, the coefficient of determination of the model was calculated based on the method developed by \citet{nakagawa2013general} using the MuMIn library in R \cite{mumin}. Approximately 14\% of the variance in the GPT-2’s subject-preference score could be explained by the fixed effects alone (marginal R2 = .142) while 23\% of the variance in the subject-preference score could be explained by both fixed and random effects (conditional $R^2 = .230$).}

\textit{Verb frequency and model performance.} We then investigated whether verb frequency is positively correlated with the language model performance. For more frequent verbs, we hypothesize that the model more closely matches the human subject/object bias. To test this hypothesis, we regressed the squared errors from the previous mixed effects model to the log-transformed word frequencies of verbs used in the stimuli. As a proxy for verb frequency in the training corpus, we used the SUBTLEX-US word frequencies from \citet{brysbaert2009moving}. We found a significant fixed effect of log-transformed word frequencies of the verbs in our materials ($b = -0.049, \textrm{SE} = 0.012, p < .01$). This supports our hypothesis: More data leads to a better approximation of human IC bias, in terms of the preference for either male or female pronouns.

\section{Study 2: Assessing continuations}\label{sec:study2}\label{sec:humanratings}

We now turn to continuations generated by GPT-2 based on the prompts described in Section~\ref{sec:prompts}. Our results above suggest that GPT-2 can generate continuations according to the causality bias pattern, given enough data. Based on this result, we now hypothesize that such continuations are better when the IC pattern is clearly present in the data. The stronger the IC bias is, the clearer the pattern of continuations in the training set. Of course, this task is much more difficult than simply generating the right pronouns. We now want to see whether the continuations actually \emph{make sense} in the eyes of human judges. To this end, we collected human ratings for a carefully controlled subset of the continuations generated by GPT-2.\footnote{For a detailed description of the full set of continuations, see Appendix~\ref{app:continuation-details}.} IRB approval was obtained prior to this study. See Appendix~\ref{app:ethicalconsiderations} for our ethical considerations.

\paragraph{Participants} We used the Prolific\footnote{\url{https://www.prolific.co}} participant pool to recruit 75 participants for the sentence rating task. Our items were spread across 25 different lists, and each participant was only allowed to provide ratings for one list. We restricted potential participants to native speakers of English, either from the UK or from the USA. After assessing response quality, we recruited five additional participants, to obtain three reliable judgments per item.

\paragraph{Task and target construct} Each continuation was assessed by three different participants, who judged whether the continuations were reasonable, given the prompt. We set up our experiment as a rating task, where each participant indicated for a list of 40 items whether they agreed with the statement that the continuation was `reasonable.' Participants could indicate their agreement on a five-point Likert scale, ranging from `Strongly Agree' to `Strongly Disagree.' With the addition of some examples in our task description (see Appendix~\ref{app:taskinstructions}), we targeted our participants' intuitions for what makes a good reason to do something. As we will discuss below, we aimed to avoid any influence of the form of the output as much as possible.\footnote{Given the terminological confusion in the field of Natural Language Generation \cite{howcroft-etal-2020-twenty}, \citet{belz-etal-2020-disentangling} developed a categorization system for evaluation criteria in NLG. In terms of their taxonomy, our informal notion of `reasonable continuation' clearly focuses on the content of the output, but the frame of reference is harder to define. If we look at the completed sentence as a whole, the sentence is evaluated in its own right, and so it is a question of \emph{Coherence}.} 

\paragraph{Prompt selection} 
Due to financial limitations, we were not able to obtain ratings for all 6888 continuations. Following the recommendations from \citet{van-miltenburg-etal-2021-underreporting}, we used a stratified sampling approach. We selected the 5 most frequent noun pairs, and the 10 most frequent verbs for each of the 10 different bias levels (as illustrated in Figure~\ref{fig:heatmaptable}). This selection gives us a sense of the upper bound performance with respect to continuation quality. Frequency was again determined using the SUBTLEX-US data \cite{brysbaert2009moving}. Prompts were constructed in the same way as before (see Ex.~\ref{ex:prompt}), with each noun pair being presented in both orders. This yields 10 verbs × 10 bias levels × 5 nouns × 2 orders = 1000 prompts.

\paragraph{Data preparation} We used GPT-2 to generate continuations for each prompt. As shown earlier, the problem with these continuations is that they may be offensive or contain disfluencies (most notably repetition, see \citealt{Fu_Lam_So_Shi_2021}). This creates two problems: (i) offensive output may cause psychological harm for our participants, (ii) offensiveness and disfluencies may lower the reliability of the ratings, if participants consistently provide lower scores for offensive/disfluent outputs (even though they may be consistent with the prompt). To prevent harm, and to avoid noise in the ratings, we took the following approach:

\begin{enumerate}[noitemsep,topsep=0pt]
    \item If the output is offensive, select a different noun pair from the 9 remaining pairs to generate a non-offensive alternative continuation.
    \item If the output contains repetition, manually remove repeated elements from the sentence, so that the core content of the continuation remains largely unchanged.
\end{enumerate}

\paragraph{Reliability} To assess annotator reliability, we used a leave-one-out approach to correlate each participant's scores with the mean scores of the two other participants who rated the same responses. Initial correlations ranged between 0.17 and 0.84. Five annotators scored below our cutoff of 0.4, and thus we recruited five more participants. After re-computing the reliability scores, we
kept only the three highest-scoring participants per task. This way, we obtained a mean score of 0.64, with a standard deviation of 0.11. This is a strong correlation, considering the subjective nature of the task.

\begin{table}
    \centering\small
    \begin{tabular}{lrrrrr}
    \toprule
    Rating & 1 & 2 & 3 & 4 & 5 \\\midrule
    Raw counts & 689 & 571 & 333 & 782 & 625 \\
    \%\thinspace of ratings & 23 & 19 & 11 & 26 & 21 \\
    \%\thinspace avg rating >= rating\hspace{-5px} & 100 & 79 & 55 & 32 & 7 \\
    \bottomrule
    \end{tabular}
    \caption{Distribution of the ratings. Ratings correspond to a Likert scale, where 1=Strongly disagree, 2=Somewhat disagree, 3=Neither agree nor disagree, 4=Somewhat agree, 5=Strongly agree.}
    \label{tab:scores}
\end{table}

\paragraph{Continuation quality} Table~\ref{tab:scores} shows the ratings in three different ways. The top row shows the (bimodal) raw score distribution: ratings tend to be either negative or positive, with relatively few ratings in the middle of the scale. The second row provides the same values as percentages, as a guide for the third row. The third row shows the percentage of continuations for which the average rating is greater than or equal to a given rating. For example: only 32\% of all the continuations have an average rating greater than or equal to 4.

A superficial analysis of the lowest and highest rated sentences suggests that the lowest ratings are often given to sentences in which swapping subject and object would improve the continuation. For example (see Appendix~\ref{app:lemons} for more):

\ex
The woman surprised the man because he was wearing a black suit and a black tie.
\xe

Future research could look into carrying out a more systematic error analysis, to create a taxonomy and present the distribution of the different kinds of errors in the continuations generated by the model \cite{van-miltenburg-etal-2021-underreporting}.

\paragraph{Explaining model performance}
We again used a mixed effects model to analyze our results. Our aim is to explain GPT-2's performance (i.e.,\ how reasonable the continuation is) in terms of verb frequency and absolute IC bias (which only looks at strength of the bias).\footnote{IC bias and log-transformed frequency values were determined as for Study 1. IC bias scores were made absolute; the mean absolute IC bias is 49.3 (SD 27.5, range [2, 92]). The mean (untransformed) frequency per million is 45.4 (SD 90.7, range [0.02, 502.27]). The correlation between IC bias and frequency was not significant and low (Pearson's $r = -0.02, p = .35$).} Full details about the model and model fitting are provided in Appendix~\ref{app:study2-statistics}.

Subject gender and its interactions were considered, but did not significantly improve model fit, and were dropped. Absolute IC bias has a small but positive effect on the rating ($ b = 0.003, \textrm{SE} = 0.001, 
p = 0.015 $). The effect of frequency is negative ($b = -0.121, \textrm{SE} = 0.038, p = 0.002$). There is also a negative interaction of bias and frequency ($b = -0.006, \textrm{SE} = 0.001, p < .001$), indicating that the positive effect of bias diminishes for higher frequency verbs. Thus, while we concluded in Study 1 that the accuracy of GPT-2 with respect to subject/object bias increases as the verb becomes more frequent, we now find that higher frequency does not give us more reasonable continuations.

\section{Discussion}
The difference between Study 1 and Study 2 indicates a qualitative difference between generating pronouns and providing explanations: The latter requires higher-level reasoning which may not be present in language models like GPT-2 \citep{bender-koller-2020-climbing}. Though the fact that GPT-2 follows the implicit causality bias in pronoun selection is at least compatible with knowledge of causality, the quality of the continuations suggests the pronoun selection is based on superficial heuristics rather than a deep understanding of language (also discussed as \emph{fast} versus \emph{slow}; see \citealt{choudhury2022machine,kahneman2011thinking}). Although existing suites for LM evaluation (e.g.\ \citealt{ettinger-2020-bert}) are useful, slower forms of assessment (such as human evaluation) are helpful to tease out this difference.

\section{Conclusion}
This paper showed two different ways to assess the linguistic capacity of a language model (GPT-2), with respect to one particular cognitively meaningful pattern, namely implicit causality. The techniques used above can be applied in a black box setting, without the need to look at the internals of the model. We hope that this paper is useful for others aiming to assess the ability of other language models to capture different linguistic phenomena as well. Our findings also showed that automatic assessment methods may not be enough to determine whether semantic phenomena like implicit causality are learned by a language model. Human evaluation remains a necessary complement to automatic evaluation \cite{VANDERLEE2021101151}. Our paper shows one way to do this without participants being influenced by factors like grammaticality and offensiveness of the output.\footnote{All code and data from this paper is available at REDACTED FOR REVIEW.}

\section{Ethical considerations}\label{app:ethicalconsiderations}
Because our study deals with human subjects, we first obtained ethical approval from our IRB. We describe our considerations below.

\subsection{Information letter and informed consent}
Our IRB mandates the use of a separate information letter (Appendix~\ref{app:informationletter}) and informed consent form (Appendix~\ref{app:informedconsent}). With the information letter, we give a general description of the study, and provide an indication of potential risks and benefits of the study. The informed consent form is provided separately to prevent information overload, and to ensure that participants know what they are agreeing to, if they decide to take part in our study.

\subsection{Crowdsourcing and payment}
Crowdsourcing has been criticized for its potentially exploitative nature \cite{fort-etal-2011-last}. We explicitly frame our task as an \emph{experiment} with \emph{human participants}, rather than a \emph{human intelligence task} with \emph{crowdworkers}, and apply the same considerations and protections as for lab experiments. Nevertheless, it is still work, and work needs to be paid. Based on experience, we expected participants to spend roughly 15 minutes on the task, and set the compensation to \textsterling 2.40, which amounts to \textsterling 9.60/hour; 10 cents above the current UK minimum wage.\footnote{See \url{https://www.gov.uk/government/publications/the-national-minimum-wage-in-2022}} In the end, the vast majority of our participants spent less time than expected on our task (range:, mean:, standard deviation:). All participants were compensated for their time, including those providing low-quality responses.

\subsection{Offensive material}
We wanted to avoid confronting our participants with profanity or otherwise potentially harmful language. We manually identified potentially offensive continuations generated by the model, and replaced harmful outputs with alternative continuations generated for different prompts. We considered continuations potentially harmful if they contained profanity or made reference to religion, violence, or sexual acts. All originally generated sentences and their replacements can be found in the GitHub repository associated with our paper.

\subsection{Language model-related harms}
Language models are associated with several different harms \cite{10.1145/3442188.3445922,DBLP:journals/corr/abs-2112-04359}, but these harms also depend on the task at hand. For example, since we used a pretrained model, our study did not incur any additional training costs. And as described above, since no one other than the authors were directly exposed to the model's output, we could prevent our participants from seeing harmful or toxic content. Thus we are mostly left with inference costs, which are relatively low, since GPT-2 can run on a personal computer.

\subsection{Intended use of this work}
This study serves two purposes: (i) to explore the ability of a language model (GPT-2) to capture native speakers' intuitions about implicit causality, and (ii) to develop an evaluation methodology that isolates coherence of the responses from other factors like offensiveness and (un)wellformedness. We do not wish to make any claims about the cognitive capacity of language models in general, nor do we want to claim that GPT-2 can somehow reason about the world. We just want to see whether the model can generate outputs that follow earlier observations about implicit causality, and that are internally consistent. Follow-up studies in the spirit of this work are encouraged.

\subsection{Licensing}
All resources used for this study were developed for research purposes, but not all materials have a clearly indicated license. GPT-2 is provided under the MIT license.\footnote{https://huggingface.co/gpt2} The work by \citet{davis-van-schijndel-2020-discourse,davis-etal-2021-computational} is provided on GitHub without any license\footnote{\url{https://github.com/forrestdavis/ImplicitCausality}} and both \citet{brysbaert2009moving} and \citet{ferstl2011implicit} published their work in the \emph{Behavior Research Methods} journal without a clear license, but with a clear intention for their work to be used for research purposes. We thus conclude that academic use of these resources is warranted.

\bibliography{anthology,custom}
\clearpage
\appendix

\section{Limitations}\label{app:limitations}
The main limitation of our paper is that we focused on only one language model (GPT-2), and only in one language (English). So while our findings provide insights into the capacities of the English GPT-2, they cannot be generalised to other language models or other languages (which is also illustrated by \citet{davis-van-schijndel-2021-uncovering}). Our main contribution is methodological, namely exploring how to assess the linguistic capacities of language models. Because our approach treats GPT-2 (mostly) as a black box, our analysis can easily be applied to other models as well. For other models, different results may be obtained.

For Study 2, a negative effect of verb frequency on the quality of generated continuations was found. As the materials used for the rating study were not gathered to cover the full range of frequencies, this pattern should not be generalized and may reflect a hidden effect. The negative interaction with IC bias suggests the same. The model fit shows an imperfect fit, but without one clear deviation from linearity. Hence, our findings may well be explained better with more appropriate independent variables.

\section{Study 1: Statistical analysis}\label{app:study1-statistics}
As noted in Section~\ref{sec:study1}, we used the lme4 \cite{lme4} and lmerTest packages in R \cite{JSSv082i13} to carry out a mixed effects regression analysis of the relationship between the model’s subject-preference score and human bias score of every IC verb, reducing the complexity by removing terms that do not significantly improve fit.

As fixed effects, human bias score, subject gender of the stimulus sentence-fragments (with two levels, namely male or female) and their interaction were included. Moreover, we included item, which indicated the pairs of antonymous nouns used in the stimulus sentences as subjects and objects, as a random effect, and added by-item random slopes for the effects of human bias score, subject gender and their interaction. However, the fitting of the full model including all fixed and random effects failed to converge. Therefore, the by-item random slopes for the effects of human bias, subject gender and their interaction were removed. Our final model is the following in R notation:

\ex
    subject-preference score \textasciitilde\ human IC bias score * subject gender + (1|item)
\xe

We regressed the squared errors obtained from the linear mixed effects regression model we performed in the previous step on the log-transformed word frequencies of IC verbs used in the stimuli. As random effects, item was also entered into the model, and by-item random slope for the effect of log- transformed frequency was added. 

\section{Study 2: Continuations}\label{app:continuation-details}
While Study 1 looked into the surprisal values for the generation of male/female pronouns, this study looks at continuations themselves. 

\subsection{Consistency check}
\begin{figure*}[htb]
    \centering
    \includegraphics[width=0.8\linewidth]{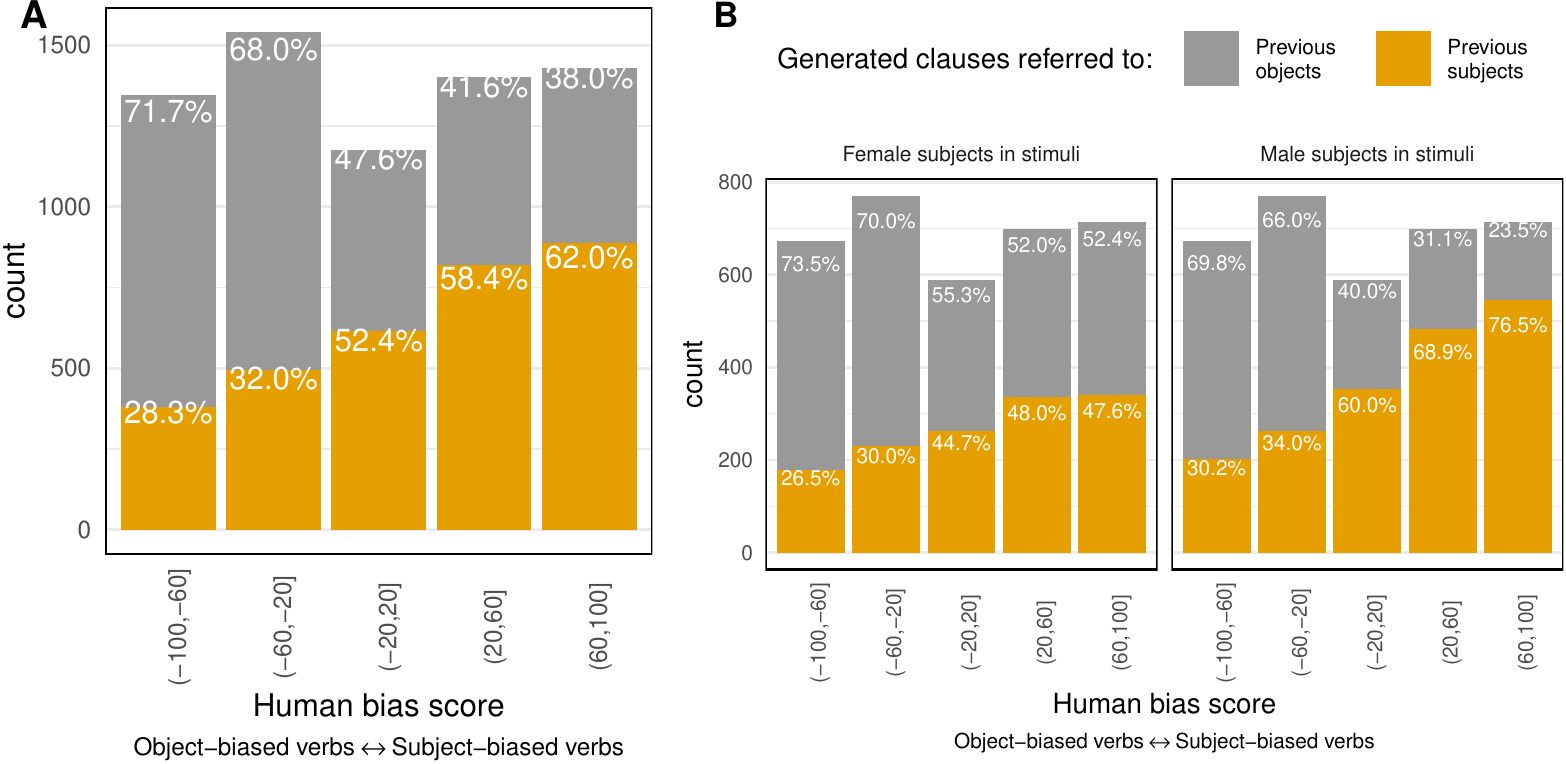}
    \caption{Percentage of continuations referring to subjects or objects by bias scores of verbs. Panel A shows overall proportion, Panel B shows proportion split by subject gender}
    \label{fig:study2reference}
\end{figure*}

We first checked whether the IC bias pattern continues to be present in the full continuations. In all but 176 cases, the model generated a gendered pronoun. These cases were coded manually, to determine whether the continuation referred to the subject or the object. Figure~\ref{fig:study2reference} shows the proportion of continuations referring to either the subject or the object of the prompt (Y-axis), split by the bias score (X-axis). We see the same trends as in the first study: overall we see that references to the subject increase as the bias score increases. This trend also holds when we split the prompts by subject gender, but for the female subjects the proportion of references to the subject never exceeds 50\%.

\subsection{Patterns} 
We then inspected the frequencies of the continuations. We observe that more than 20\% of the outputs is repeated more than 100 times. This lack of diversity is a common issue in (neural) Natural Language Generation (e.g.\ \citealt{van-miltenburg-etal-2018-measuring,hashimoto-etal-2019-unifying,zhang-etal-2021-trading}). It is not necessarily a problem at the individual level (the continuation may be a bit generic but still appropriate for the given context), but at the corpus level these `one size fits all' continuations are shortcuts that prevent more varied outputs.\footnote{We might also question whether it is possible at all to properly assess the cognitive capacity of a model that keeps using such shortcuts (which may be seen as cheating). As an alternative, we can imagine a two-stage process where researchers first generate an unrestricted set of continuations, and then force the model to avoid common continuations.} 

\subsection{Offensive output}
Looking over the generated outputs, there were several occasions where the model generated offensive continuations, including instances of sexism, racism, and misogyny. On top of this, the model outputs also contained sexually explicit words, and some continuations described acts of violence. Following the recommendations from \citet{2022arXiv220414256D}, we do not provide any examples in this paper. As discussed in Section~\ref{sec:humanratings}, we also removed (potentially) offensive continuations from our human rating experiment. For transparency reasons, we do provide those sentences in our GitHub repository.

\begin{table*}
    \centering\small
    \begin{tabular}{lrclr}
\toprule
\multicolumn{2}{c}{\textbf{Male subject}} & & \multicolumn{2}{c}{\textbf{Female subject}}\\
Continuations & Frequency & & Continuations & Frequency\\
	\midrule
he was afraid of her & 521 & &	she was afraid of him. &	476\\
she was a woman &	102 & & he was a good man and he was a good man &	166\\
he was a good man and he was a good man &	68 & &he was a good man &	124\\
she was a good girl &	64 & &he was a good boy &	92\\
she was a woman, and he was a man &	60 & &	she was a woman &	57\\
\bottomrule
    \end{tabular}
    \caption{Most frequent continuations for the prompts in Study 2, split by subject.}
    \label{tab:frequentcontinuations}
\end{table*}

\subsection{Repetitions}\label{app:study2-results}
Table~\ref{tab:frequentcontinuations} shows the five most frequent continuations for our prompts, split by subject gender. It is clear from the table that the continuations generated by GPT-2 are very repetitive, and tend to be generic without any specific details.

\section{Study 2: Statistical analysis}\label{app:study2-statistics}
A full linear model was built with the factors subject gender, absolute IC bias (centered), log-transformed word frequency (z-scored) and their interactions as fixed effects, and random effects for participant (intercept, and slopes for the three main effects), sentence (intercept), and subject/object pair (intercept). The model was then subjected to the step function of lmerTest, which removes insignificant components. Though the full model and some of the first reductions of it did not converge, this procedure is still appropriate (a model that does not converge should not be used, and as more data is not available, the model should be simplified). The reduced final model, that did converge, is, in R syntax:

\ex
    rating \textasciitilde\ abs(IC bias\_abs\_c) * frequency + (1 | sentence\_ID) + (1|participant)  + (1|item)
\xe

The DHARMa package \cite{DHARMa} was used to assess model fit. Though the residuals are not normally distributed, the deviations show no clear pattern. To avoid spurious conclusions, we corroborated the significant estimates by checking if their 95\% confidence intervals included 0, which they did not. Hence, treating ratings as a continuous numeric variable was not problematic.

\begin{figure}[htp]
\includegraphics[width=\linewidth]{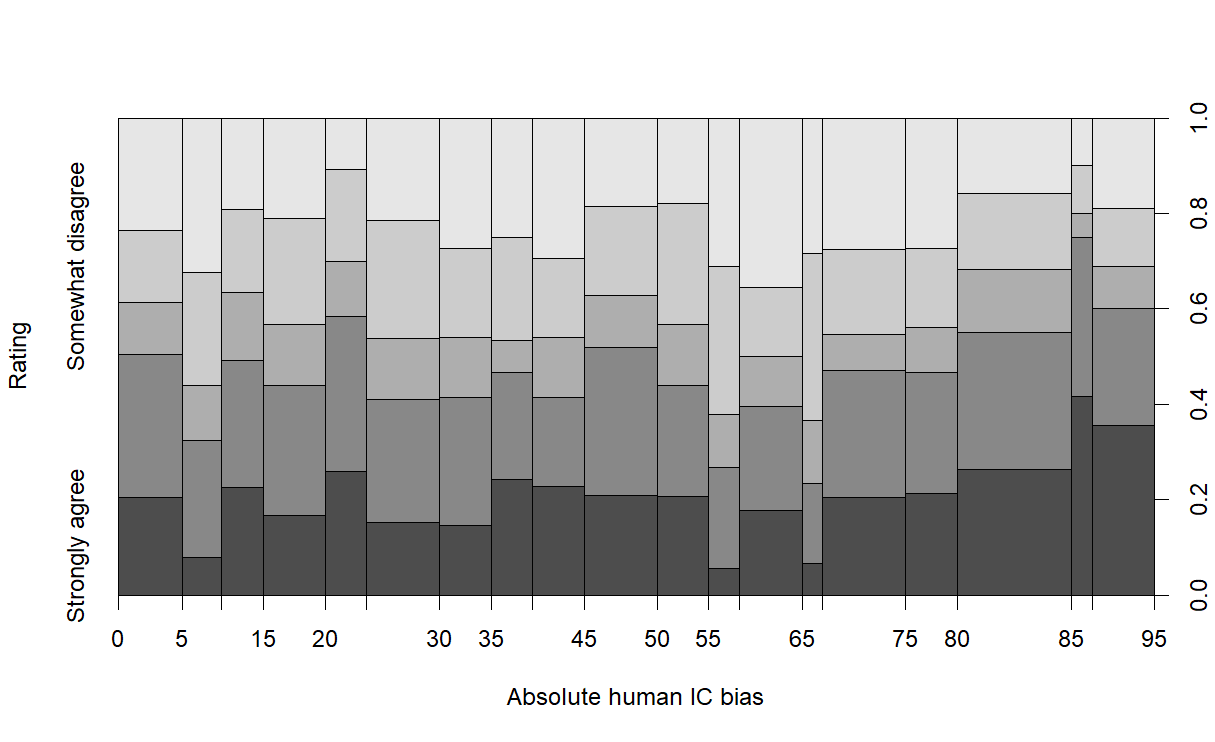}
\caption{Ratings by absolute value of human IC bias.\label{fig:study2bias}}
\end{figure}

\begin{figure}[htp]
\includegraphics[width=\linewidth]{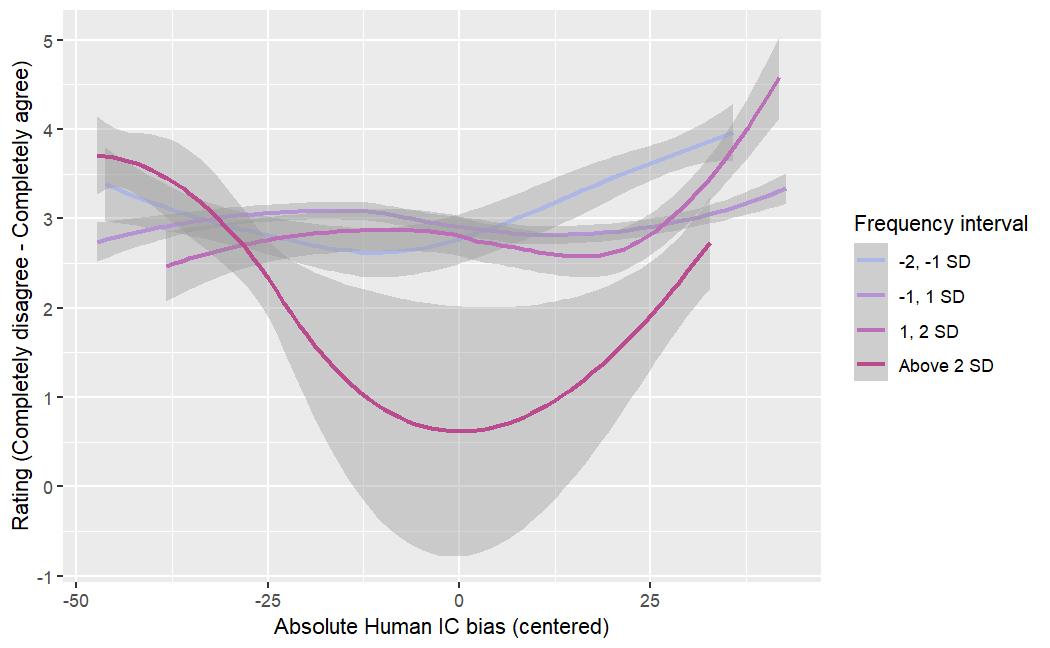}
\caption{Ratings by absolute value of human IC bias, separated for four different levels of verb frequency.\label{fig:study2}}
\end{figure}

\section{Study 2: Details about the ratings}
\label{app:lemons}
A total of 72 sentences was only rated at the lowest level. Three random examples are:

\begin{enumerate}[noitemsep]
    \item The man fed the woman because he was afraid of her.
    \item The woman worried the man because he was a bigger man. \item The woman surprised the man because he was wearing a black suit and a black tie.
\end{enumerate}

Tentatively, we interpret these sentences as a mix-up of two patterns. The first sentence continues with the subject, but gives no good explanation (in fact, being afraid of someone would not make feeding them a good idea). The second and third example continue with an object explanation, but would make more sense if they had continued with a subject explanation, i.e., flipping the gender, e.g., `The woman surprised the man because \emph{she} was wearing \dots'.

There were 69 continuations that always received the highest rating tentatively seem to conform to the human IC bias. One random example is `The woman thanked the man because he was a good man.`; the verb `to thank' is strongly object-biased (raw subject bias score -92).

\section{Information letter}\label{app:informationletter}
\textbf{Title}: Assessing computer-generated texts\\

\noindent\textbf{Introduction}\\
We invite you to take part in our study to assess computer-generated texts. This study is part of a larger project to see how good or bad computers are at producing or understanding human language,such as English. In this study you will be asked to rate the quality of computer-generated texts/sentences. We will use this information to see what the computer is good at, and to see where it can still be improved.\\

\noindent\textbf{What do I have to do?}\\
As mentioned above, you will be asked to rate the quality of computer-generated texts. In this study, you will be asked to read 40 short sentences, and to provide your judgment. We are interested in your intuition as a native speaker of English, so you don’t need to think too long about it.\\

\noindent\textbf{Expected duration}\\
We expect this study to take about fifteen minutes of your time. Other than this, we do not foresee any risks associated with this study. On the positive side, your participation will improve our understanding of the language capacity of modern computer models.\\

\noindent\textbf{Ethics and rights}\\
This study was approved by the “Research Ethics and Data Management Committee” of REDACTED.\\

Your participation is completely voluntary. Your consent to participate generally applies for the duration of this study. However, you have the right to decline to participate and withdraw from the research once participation has begun, without any negative consequences, and without providing any explanation.\\

Your participation is completely anonymous. We will not store any identifying information, so your answers cannot be traced back to you. We only see the demographic information that Prolific provides. Do let us know if you would like to have a copy of your responses, and we will try to obtain them based on the Prolific ID.\\

\noindent\textbf{Use of data}\\
Your responses will be used for the current study, and possible follow-up studies in the future. This means that the data will be presented in research articles, that are publicly available. For full transparency, we will also publicly share the anonymised individual responses. As such, they will be stored indefinitely.\\

\noindent\textbf{Contact}\\
If you have any questions, or if would like to learn more about this study, please contact REDACTED for any of your questions.\\

\noindent\textbf{Ethics approval}\\
If you have any remarks or complaints regarding this research, you may also contact the “Research Ethics and Data Management Committee” of REDACTED via REDACTED.

\section{Informed consent}\label{app:informedconsent}
By agreeing to this consent form, you confirm that you have read the study description and that you have been offered the opportunity to ask questions (via email). Remember that your participation is voluntary, and that you have the right to decline to participate and withdraw from the research once participation has begun, without any negative consequences, and without providing any explanation.\\

\noindent I hereby give permission to:
\begin{itemize}[topsep=0pt,noitemsep]
    \item Store my anonymised responses to this survey.
    \item Analyse the anonymised data (both manually and automatically through statistical software).
    \item Make the responses to this survey publicly available upon completion of the study.
\end{itemize}

\null

\noindent Yes {\color{gray}$\Rightarrow$ continue to the survey.}\\
No {\color{gray}$\Rightarrow$ continue to the end of the survey.}

\section{Task instructions}\label{app:taskinstructions}
\textbf{Task instructions}\\
All questions below are of have the same form. You will see the start of a sentence on the first line, and a continuation generated by a computer model on the second line. Your job is to assess the quality of the continuation on the second line.\\

\noindent \textbf{Example of a reasonable continuation:}\\
For the following sentence: \textit{The clown startled the girl because}\\
A reasonable continuation would be: \textit{his make-up was scary.}
\begin{itemize}[noitemsep,topsep=0pt]
    \item Strongly agree
    \item Somewhat agree
    \item Neither agree nor disagree
    \item Somewhat disagree
    \item Strongly disagree
\end{itemize}

\noindent This is a reasonable continuation because scary make-up can cause someone to be startled. So here you would answer \textit{Strongly agree}.\\

\noindent \textbf{Example of a less reasonable continuation:}\\
For the following sentence: \textit{The clown startled the girl because}\\
A reasonable continuation would be: \textit{she liked him.}

\begin{itemize}[noitemsep,topsep=0pt]
    \item Strongly agree
    \item Somewhat agree
    \item Neither agree nor disagree
    \item Somewhat disagree
    \item Strongly disagree
\end{itemize}

This is a less reasonable continuation, because being liked by someone is generally not a reason to startle them. So here you would answer one of the disagree options.

\end{document}